\documentclass[10pt,twocolumn,letterpaper]{article}

\usepackage{wacv}
\usepackage{times}
\usepackage{epsfig}
\usepackage{graphicx}
\usepackage{amsmath}
\usepackage{amssymb}
\usepackage{booktabs}
\usepackage{algorithm}
\usepackage{algorithmic}

\wacvfinalcopy  
\ifwacvfinal
\usepackage[breaklinks=true,bookmarks=false]{hyperref}
\else
\usepackage[pagebackref=true,breaklinks=true,colorlinks,bookmarks=false]{hyperref}
\fi

\begin{document}

\title{ROMA: Run-Time Object Detection To Maximize Real-Time Accuracy}

\author{JunKyu Lee\\
Queen's University Belfast\\
Belfast, UK\\
{\tt\small junkyu.lee@qub.ac.uk}
\and
Blesson Varghese\\
University of St Andrews\\
St Andrews, UK\\
{\tt\small blesson@st-andrews.ac.uk}
\and
Hans Vandierendonck\\
Queen's University Belfast\\
Belfast, UK\\
{\tt\small h.vandierendonck@qub.ac.uk}
}

\maketitle

\begin{abstract}
This paper analyzes the effects of dynamically varying video contents and detection latency on the real-time detection accuracy of a detector and proposes a new run-time accuracy variation model, ROMA, based on the findings from the analysis. ROMA is designed to select an optimal detector out of a set of detectors in real time without label information to maximize real-time object detection accuracy. ROMA utilizing four YOLOv4 detectors on an NVIDIA Jetson Nano shows real-time accuracy improvements by 4 to 37\% for a scenario of dynamically varying video contents and detection latency consisting of MOT17Det and MOT20Det datasets, compared to individual YOLOv4 detectors and two state-of-the-art runtime techniques. 
\end{abstract}

\section{Introduction}

Real-time object detection plays a fundamental role in various applications such as self-driving cars, real-time object tracking, real-time activity recognition, and robotics~\cite{towards-wang, lee-tod, korshunov-reducing}.
Obtaining high real-time detection accuracy is often challenging due to domain shift issues \cite{challen-artificial, filos-can, ganin-unsupervised, bousmalis-domain} and real-time latency constraints~\cite{korshunov-reducing}. Several domain adaptation techniques improved the generalization ability of the object detection models by adapting the feature representation of a backbone Deep Neural Network (DNN) to massive unlabeled data~\cite{ganin-unsupervised, bousmalis-domain, filos-can}.   
 
However, using a single detector is limited in improving detection performance on dynamically varying video contents and dynamically varying compute resources, due to a fixed backbone network. In this regard, many run-time techniques explored how to select an optimal selector out of multiple detectors to improve the detection performance~\cite{lee-tod} or to save compute resources, given an accuracy budget~\cite{jiang-chameleon, lou-dynamic-ofa}. 
\Eg, switching between multiple detectors in real-time according to object size distribution in test data can improve the detection performance, compared to utilizing a single detector \cite{lee-tod}.    

Previous approaches utilizing multiple detectors~\cite{lee-tod, jiang-chameleon, lou-dynamic-ofa} are limited in practice for real-time object detection applications. \Eg, a run-time approach selects an optimal network based on periodic accuracy assessment, requiring labels from data~ \cite{jiang-chameleon}. However, the ground truths are not known in real-time for many real-time applications. Another run-time approach selects an optimal network assuming that available compute resources are fixed~\cite{lee-tod}. However, the available compute resources can vary in practice according to the background workload. Lou et al.~\cite{lou-dynamic-ofa} considered dynamically varying compute resources but did not consider the impact of dynamically varying object sizes and speeds on the accuracy. Therefore, a natural question arises: how can an optimal detector be selected without accessing labeled data based on the effects of both dynamically varying video contents and available compute resources? To the best of our knowledge, no solution to this problem exists in the literature. 


Real-time accuracy highly depends on the available compute resources, unlike offline detection accuracy. \Eg, if a computing device is shared to analyze multiple video streams, adding a new video analysis task will reduce the amount of computation available to the other tasks. The reduced compute resources increase the detection latency, degrading the real-time performance. The increased latency can be addressed by dropping frames~\cite{lee-tod, korshunov-reducing} or downsampling frames~\cite{jiang-chameleon, lee-yolo}. 
As such, the computational efficiency of real-time video analytics is tightly bound to its real-time accuracy. 

We model the effects of dynamically varying object sizes, moving speeds, and detection latency on the real-time accuracy of each detector by separating the effects into two parts: the effects of dynamically varying objects on the accuracy and the effects of dynamically varying object speeds and latency on the accuracy. 
We establish a new model based on this idea 
and this model can be used to choose the best performing detector out of multiple detectors without label information. 
The model estimates are computed based on information available at run-time: characteristics of the objects detected in the current and previous frame, and the detection latency. We name our run-time accuracy estimation model as $\underbar{R}$untime $\underbar{O}$bject Detection Accuracy Variation Estimation to $\underbar{M}$aximize Real-Time $\underbar{A}$ccuracy (ROMA). 
The main contributions of this paper include:
\begin{itemize}
    \item Analysis of the effects of dynamically varying compute resources and video contents on the real-time accuracy variation. 
    \item A novel run-time accuracy estimation model, ROMA, that estimates the Relative Average Precision (RAP) of each detector to a currently running detector without label information. ROMA is designed, independently of detector types and computing platforms. 
    \item Demonstrating ROMA using multiple YOLOv4 detectors \cite{bochkovskiy-yolov4} on an NVIDIA Jetson Nano with the Multiple Object Tracking Challenge 2017 (MOT17Det) and 2020 (MOT20Det) datasets \cite{MOTDet} for dynamically varying video content and compute resources use cases. 
\end{itemize}
We present related work in section~\ref{sec:related work}, analysis of real-time accuracy variation and the implementation of ROMA in section~\ref{sec:analysis}, the experimental evaluation in section~\ref{sec:experiment}, and conclude our paper in section~\ref{sec:conclusion}.

\section{Related Work} \label{sec:related work}





Several researchers attempted to control the frame rate of video streams at run-time according to dynamically varying video contents to improve real-time detection accuracy \cite{jiang-chameleon, lee-yolo, korshunov-reducing}. Korshunov et al.~\cite{korshunov-reducing} discussed that slower-moving objects were correctly detected when a higher fraction of the frames were dropped, implying that high frame rates were unnecessary in such cases.
Mohan et al.~\cite{mohan-determining} estimated the least sufficient Frame Per Second (FPS) during run-time for object tracking applications to save bandwidth between cameras and the compute devices, given the accuracy budgets. 
\Eg, if the tracked object is within a distance threshold on the subsequent frames, the FPS is lowered until it violates the threshold. 

Zoph et al.~\cite{zoph-neural} proposed Neural Architectural Search (NAS) to seek optimal DNN models in the space of hyperparameters of network width, depth, and resolution. Since then, many NAS variants attempted to seek resource-efficient DNNs to deploy them on resource-constrained devices \cite{he-amc, Tan-mnasnet, tan-efficientnet, tan-efficientdet, wu-fbnet, anderson-performance, cai-once}. 
\Eg, Tan et al.~\cite{tan-efficientdet} proposed a resource-aware NAS, \textit{Efficientdet}, to seek resource-efficient detectors for object detection applications. 
Recently, a feed-forward NAS approach~\cite{cai-once} produced resource-efficient DNNs, given computing resource and latency constraints. 

A different strand of work
aims to select the most appropriate detector
from a set of available
detectors~\cite{lou-dynamic-ofa, lee-tod, jiang-chameleon}
or an appropriate channel width out of a single multi-capacity detector \cite{fang-nestdnn, yu-slimmable}.
Lee et al.~\cite{lee-tod} exploit temporal locality and select an optimal detector in real-time based on the sizes of objects in the video. Boundaries between object sizes were empirically determined
using the MOT17Det dataset~\cite{MOTDet}.
Their run-time technique ``Transprecise Object Detection (TOD)'' selects
the detector that corresponds to the median object size found in the last frame.
Lou et al.~\cite{lou-dynamic-ofa} propose a Latency-Aware Detection (LAD) run-time technique that selects the detector with the highest latency that meets the required frame rate, which corresponds to the detector with the highest off-line detection accuracy that meets the frame rate. 
Both TOD~\cite{lee-tod} and LAD~\cite{lou-dynamic-ofa} required multiple resource-efficient detectors to be uploaded to DRAM at initialization time to minimize the time overhead of switching detectors. \Eg, TOD required an 11\% additional memory footprint to upload four different detectors on an NVIDIA Jetson Nano device, compared to uploading a single heavyweight detector out of the four detectors \cite{lee-tod}. Yu et al.~\cite{yu-nisp} propose the run-time technique of pruning unimportant neurons. A lightweight run-time decision maker \cite{marco-optimizing} switched between multiple detectors during run-time to improve image classification accuracy.
Minhas et al.~\cite{minhas-leveraging} selected an appropriate detector model according to dynamically varying accuracy constraints to improve the inference throughput. 

\section{ROMA: Run-Time Accuracy Variation} \label{sec:analysis}
We discuss the real-time accuracy trade-off between the offline accuracy and the latency of a detector using the Average Precision (AP) metric. Later, we discuss the ROMA.   

\subsection{Real-Time Accuracy Characteristics}
\begin{figure*}
\begin{center}
 \includegraphics[width=\linewidth]{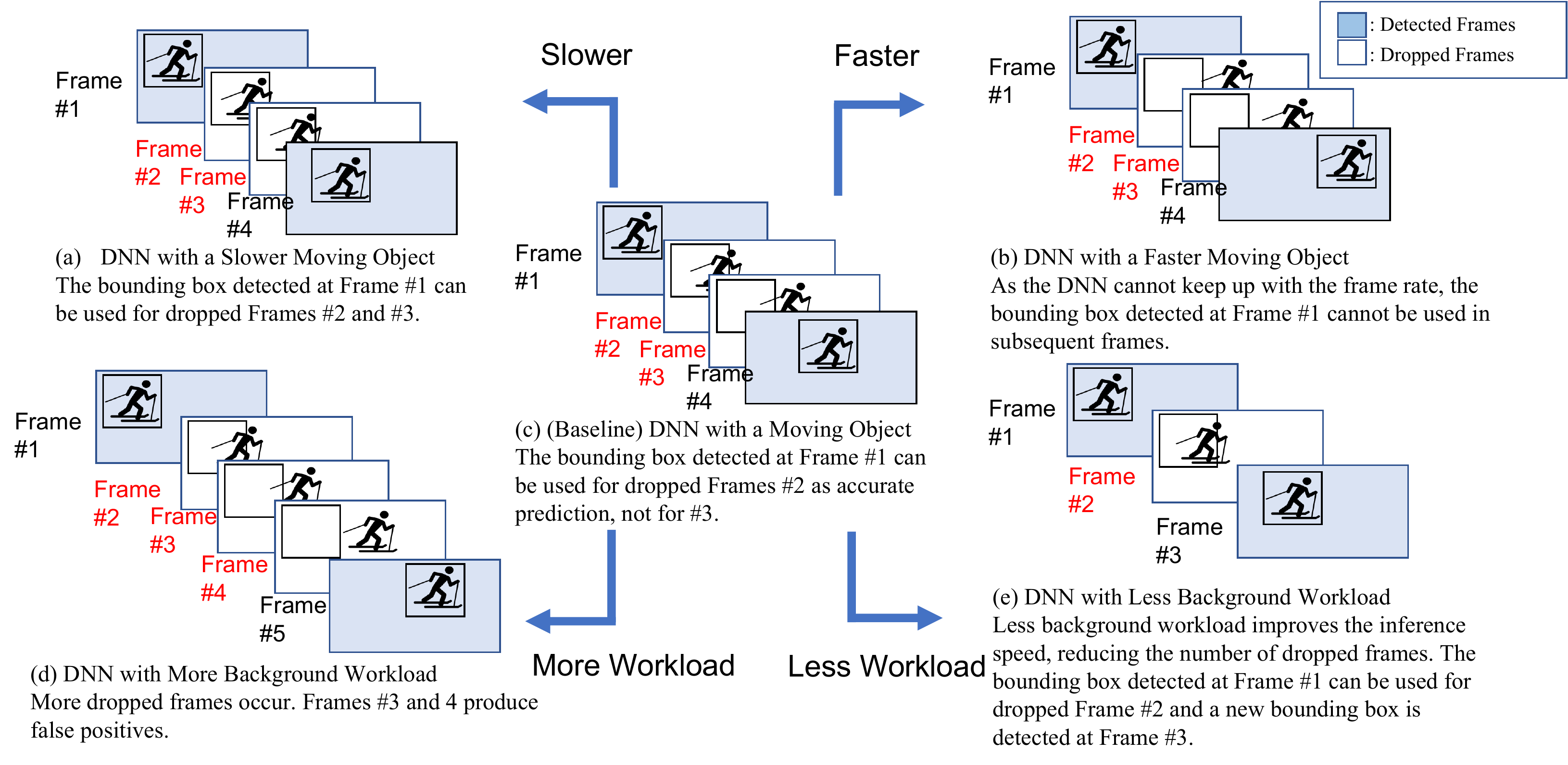}
\end{center}
  \caption{Accuracy Variation with Dynamically Varying Objects' Speeds and Available Compute Resources}
\label{fig:high_level}
\end{figure*}

Central to our approach to processing the video stream in real-time without running behind is to drop frames when the frame rate cannot
be achieved~\cite{korshunov-reducing, lee-tod}. When frames are dropped, we apply the same object detection bounding boxes as the previous frame that was analyzed. Such a substitution is meaningful as the contents of a video frame are typically not very different from the contents of the previous video frame. However, the approach is not exact, and accuracy declines as more frames are dropped.
Figure~\ref{fig:high_level} analyzes the key problems that may occur when copying the detection bounding boxes from analyzed frames to subsequently dropped frames.

Consider case (c) (Figure~\ref{fig:high_level}, center) as a baseline. As the skier moves, it leaves the bounding box identified for frame~\#1. In frame~\#2, the overlap between the previously identified bounding box and the skier is sufficient to consider a correct detection (\eg, an Intersection of Union (IoU) is larger than 50\%). However, in frame~\#3, the skier has moved further and the detection fails, based on information in frame~\#1. As such, dropping frame~\#2 still allows obtaining a correct prediction, while dropping frame~\#3 results in an incorrect prediction.

Accuracy for dropped frames is clearly dependent on the video contents, in particular the speed at which objects move and the original video frame rate.
Case~(a) shows a scenario where the movement of the skier is less than
in case~(c). In this case, the skier can be detected correctly in
both frames~\#2 and~\#3 when those frames are dropped, i.e., the bounding box found in frame~\#1 is reused. In case~(b), the speed of the skier is higher, and the bounding box found in frame \#1 quickly becomes stale in both frames \#2 and \#3.

Another major factor that impacts on real-time object detection accuracy
is computational latency. Computational latency may vary across
object detectors (trading-off complexity of the detector against its
accuracy), or when the compute hardware is shared with other workloads.
When it takes longer to analyze a frame, then
a higher number of subsequent frames need to be dropped in order to keep up
with the real-time frame rate (case~(d)). Alternatively, if frames can be
analyzed more quickly (case~(e)), a higher fraction of frames can be analyzed.
As such, one expects a higher detection accuracy.

Note that in all cases, the frame rate at which the video is also presented
impacts on detection accuracy. The proposed approach has the advantage that it
matches the number of analyzed frames to the compute capability that is
available by dropping as many frames as necessary.
Increasing the frame rate may result in all additional frames being dropped
when insufficient compute resource is available. Lowering the frame rate may
result in lower accuracy on sufficiently available compute resources.
As such, we will assume in our experiments
a fixed frame rate that is slightly higher than what
can be analyzed in real time using the hardware and object detectors
used.



In real-time object detection, the choice of object detector has
a complicated impact on accuracy, as it simultaneously impacts on several
factors. A heavyweight detector, which is more complex and thus requires
more computation to be performed, will achieve higher accuracy than a
lightweight detector~\cite{huang-speed}.
However, by requiring more computation, a higher
number of frames will need to be dropped
(Figure~\ref{fig:high_level}, case~(d)), which negatively impacts on accuracy.
Similarly, a lightweight detector may be able to analyze more frames, but it
does so with less accuracy than the heavyweight detector.

Additionally, the contents of the video frames impacts on accuracy.
\Eg, lightweight detectors achieve comparable accuracy to heavyweight detectors on large objects, but not so on small objects~\cite{huang-speed}.
The goal of this paper is to detangle the complex interaction between video content and latency of executing the object detector for real-time accuracy of an object detector.
As a use case, we exemplify one of stage-of-the-art YOLO detectors, YOLOv4~\cite{bochkovskiy-yolov4}. A YOLOv4 employs either a 9-layered DNN backbone for a tiny version~\cite{feng-tinier-yolo} or a 53-layered DNN backbone for a full version~\cite{bochkovskiy-yolov4}. The two knobs, the DNN's resolution $r_{DNN}$ and the detector structure (i.e., tiny vs full version), can be used to control the trade-off between speed and offline AP of YOLOv4 in real-time. \Eg, smaller-sized objects can be detected more by employing a higher $r_{DNN}$ or a full version detector. However, the increased detection latency can hurt real-time accuracy. 

\subsection{Notations for Frames and Frame Block Sizes}
We describe our model that estimates RAPs in this section. In the beginning, we use the notations for $n$ detector candidates in a detector pool $\mathbf{d}$ as follows: 
\begin{equation} \label{eq:detector list}
 \mathbf{d} = \{d_1, ..., d_{n}\}.    
\end{equation}

A video stream consists of a sequence of frames, which are analyzed one by one by an object detector. The $i_f^{th}$ frame in the sequence is identified as $f_{i_f}$, where $i_f =0, 1, 2,\ldots$. Some frames are analyzed by the detector, while others are dropped when frames arrive faster than they can be analyzed. The model estimates RAPs using a series of frames consisting of the detected frame and subsequently dropped frames. We refer to such a series of frames associated with a detector as \textit{a frame block} of a detector and the number of frames consisting of a frame block as \textit{a frame block size}. \Eg, the first frame in a frame block is analyzed by the object detector and the remaining ones have been dropped. Therefore, the first analyzed frame is commonly used for each frame block associated with each detector and the frame block size of each detector depends on the number of dropped frames of a detector. Thus, we use the notation $f_{s(t)}$ for the $t^{th}$ analyzed frame, where $t=0,1,2,\ldots$ and $s(t)$ associates with a corresponding frame index $i_f$. This way, we can link the analyzed frame index $t$ to a frame sequence number $i_f$.

\begin{figure}
\centering
 \includegraphics[width=\linewidth]{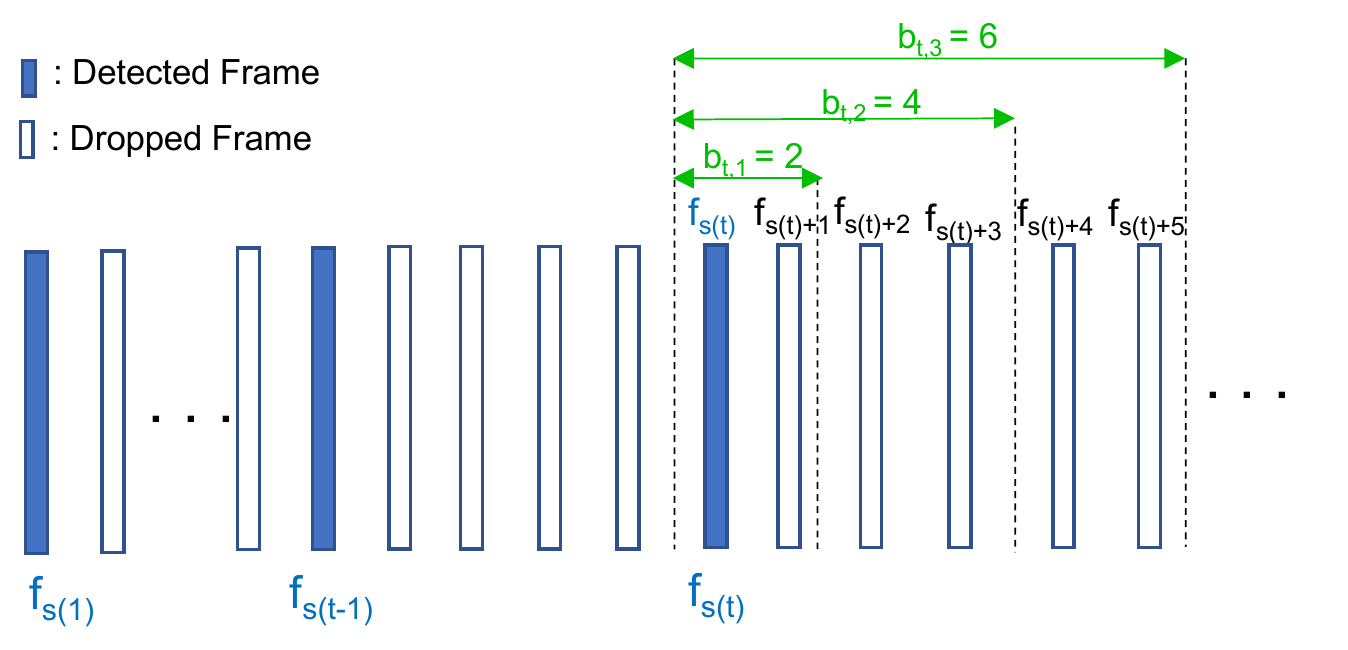}
  \caption{Notations for Frames and Frame Block Sizes}
\label{fig:frame_notations}
\end{figure}

Fig.~\ref{fig:frame_notations} describes our mathematical notations for frames and frame block sizes with an example for the number of detectors $n=3$.
As such, a frame block starting with $f_{s(t)}$ consists of frames $f_{s(t)},f_{s(t)+1},f_{s(t)+2},\ldots,f_{s(t)+b_{t,i}-1}$, where $b_{t,i}$ indicates the \textit{frame block size} of a frame block starting at $f_{s(t)}$ associated with a detector $d_i$. Likewise, we use the notation $d_{c(t)}$ for a currently chosen detector to detect objects at the frame $f_{s(t)}$, where $c(t)$ associates with a detector index $i$. We will estimate RAP of each detector $d_i$ based on \textit{its own frame block}, compared to a currently running detector $d_{c(t)}$. 

The AP at $f_{s(t)}$ is the offline AP of a detector, and the AP is expected to degrade gradually as the number of dropped frames increase in proportion to the expected number of missing objects due to IoU deviation between $f_{s(t)}$ and a dropped frame. {Hence, ROMA estimates the offline AP of each detector and then seeks the accuracy degradation rate at each dropped frame.}

\subsection{Offline AP Estimation of Each Detector} \label{sec:number of detected objects}
If the precision distribution is equivalent among the objects detected from each detector, the recall becomes the main factor in determining the AP. The recall is improved in proportion to the number of detected objects which highly depends on object size distribution on a video frame. 
Based on the above assumption, we estimate the number of objects detected of each detector $d_i$ for offline AP estimation using the object size distribution detected at $f_{s(t)}$ using $d_{c(t)}$. 

To do so, we first seek the detection performance ratios between $d_i$ and $d_{c(t)}$ at different object size regions using an offline dataset (i.e., not used for evaluation dataset). We divide the object sizes (i.e., the number of pixels) into the $H$ regions and measure the number of detected objects of each detector $d_i$ to form an $\mathbf{p}_{i}$ vector as follows:
\begin{equation}
    \mathbf{p}_{i} = [p_1(d_i), p_2(d_i), ... ,  p_H(d_i)]^T,
\end{equation}
where $p_k(d_i)$ is the number of detected objects at the region $k$ using the detector $d_i$. Utilizing each $\mathbf{p}_{i}$ generates a prior histogram matrix $\mathbf{P}$ as follows:
\begin{equation}
\mathbf{P} = [\mathbf{p}_1, \mathbf{p}_2, ... , \mathbf{p}_n]^T.    
\end{equation}
Next, the number of objects directly detected on $f_{s(t)}$ using the current detector $d_{c(t)}$ with respect to each region can generate a vector $\tilde{\mathbf{p}}_{t,c(t)}$ as follows:
\begin{equation} \label{eq:h_it}
\tilde{\mathbf{p}}_{t,c(t)} = [\tilde{p}_1(d_{c(t)}), \tilde{p}_2(d_{c(t)}), ... , \tilde{p}_H(d_{c(t)})],
\end{equation}
where $\tilde{p}_k(d_{c(t)})$ is the number of objects detected at region $k$ on $f_{s(t)}$ using the detector $d_{c(t)}$. 

Next, the relative number of detected objects of $d_i$ to $d_{c(t)}$ at each region is estimated as follows:
\begin{equation} \label{eq:r_it}
\mathbf{r}_{i,c(t)} =  [p_1(d_i)/ p_1(d_{c(t)}), ... , p_H(d_i)/ p_H(d_{c(t)})]^T.   
\end{equation}
Notice that $\tilde{p}_k(d_{c(t)})$ is run-time detection information which varies over time according to video contents on $f_{s(t)}$ while ${p}_k(d_{c(t)})$ is offline detection information using offline data, which is fixed over time. From this point forward, we will use tilde notations for data measured during run-time (e.g., $\tilde{\mathbf{p}}_{t,c(t)}$).

Finally, we estimate the number of objects detected on $f_{s(t)}$ using a detector $d_i$ by utilizing both the run-time information $\tilde{\mathbf{p}}_{t,c(t)}$ and the detection ratio information $\mathbf{r}_{i,c(t)}$:
\begin{equation} \label{eq:l_it}
 l_{t,i} = \mathbf{r}_{i,c(t)}^T \tilde{\mathbf{p}}_{t,c(t)}.
\end{equation}
If $i=c(t)$, we use the tilde notation, $\tilde{l}_{t,c(t)}$, since the number of detected objects at $f_{s(t)}$ is directly measured rather than estimated.  

\subsection{AP Degradation at Each Dropped Frame} \label{sec:num_objs at dropped frames}
We estimate the AP degradation rate at each $j^{th}$ dropped frame (e.g., $f_{s(t)+j}$, where $j \geq 1$), compared to the AP on the analyzed frame $f_{s(t)}$. We use the notation $AP_{t,i}(f_{s(t)+j})$ for the estimated AP of a detector $d_i$ at the frame $f_{s(t)+j}$ in a frame block and the notation $\overline{AP}_{t,i}$ for the estimated average AP of $d_i$ over frames in a frame block starting with the frame $f_{s(t)}$:
\begin{equation} \label{eq:AP general}
  \overline{AP}_{t,i} = \Sigma_{j=0}^{b_{t,i}-1} AP_{t,i}(f_{s(t)+j})/b_{t,i}.
\end{equation}

Each $AP_{t,i}(f_{s(t)+j})$ is either lower than or equal to $AP_{t,i}(f_{s(t)})$, since some of detected objects on $f_{s(t)}$ can be lost due to limited overlap between objects' bounding boxes multiple frames apart. In this regard, we model $AP_{t,i}(f_{s(t)+j})$ by introducing an accuracy degradation ratio parameter as follows:
\begin{equation} \label{eq:AP dropped frames}
    AP_{t,i}(f_{s(t)+j}) = AP_{t,i}(f_{s(t)}) \times \beta_{s(t)+j},
\end{equation}
where each $\beta_{s(t)+j}$ represents an AP degradation ratio at the frame $f_{s(t)+j}$, compared to the AP at $f_{s(t)}$ (e.g., $\beta_{s(t)} = 1$ always and $0 \leq \beta_{s(t)+j} \leq 1$, where $j \geq 1$). We notice that $\beta_{s(t)+j}$ can be shared among all detectors, since the ratio mainly relies on the average of detected objects' moving speeds depending on video contents rather than a detector type. 

We estimate $\beta_{s(t)+j}$ with three steps. In step 1, we estimate the frame block size $b_{t,i}$ of $d_i$. In step 2, we estimate the number of missing objects per dropped frame due to the IoU deviations using $b_{t,i}$. 
In step 3, $\beta_{s(t)+j}$ is estimated using the estimated number of missing objects per frame. 

For the step 1, the detection latency of $d_i$, $L_{t,i}$, determines $b_{t,i}$. 
If a detector is not switched between $f_{s(t-1)}$ and $f_{s(t)}$, $L_{t,i}$ is updated as follows: 
\begin{equation} \label{eq:L_it1}
    L_{t,i} = (\tilde{L}_{t,c(t)}/{L}_{t-1,c(t)}) \times L_{t-1,i} \text{, if $c(t) = c(t-1)$}. 
\end{equation}
This way, 
if available compute resources varies between $f_{s(t-1)}$ and $f_{s(t)}$, 
$L_{t,i}$ is updated by using a latency variation ratio $\tilde{L}_{t,c(t)}/{L}_{t-1,c(t)}$. 
If the detector is changed between $f_{s(t-1)}$ and $f_{s(t)}$ (i.e., $c(t-1) \neq c(t)$), the estimated latency $L_{t,c(t)}$ is directly updated to the measured latency $\tilde{L}_{t,c(t)}$:
\begin{equation} \label{eq:L_it2}
    L_{t,c(t)} = \tilde{L}_{t,c(t)} \text{, if $i = c(t)$ and $c(t) \neq c(t-1)$}.  
\end{equation}
The rest of estimated latency $L_{t,i}$ of the other detectors are unchanged if $c(t-1) \neq c(t)$:
\begin{equation} \label{eq:L_it2_again}
    L_{t,i} = {L}_{t-1,i} \text{ if $i \neq c(t)$ and $c(t) \neq c(t-1)$}.  
\end{equation}
We do not utilize Eq.~(\ref{eq:L_it1}) for $c(t) \neq c(t-1)$, so that the update of other detectors' latency utilizes the latency ratio derived only from the direct measurements (
i.e., $L_{t-1,c(t)} = \tilde{L}_{t-1,c(t)}$ if $c(t) = c(t-1)$). 
Using an $L_{t,i}$ and an FPS constraint $FPS$, the $b_{t,i}$ is estimated as follows: 
\begin{equation} \label{eq:b_it}
    b_{t,i} = f(FPS, L_{t,i}) = \lfloor FPS \times L_{t,i} \rfloor + 1.
\end{equation}
Notice that each $b_{t,i}$ is varying with $t$ according to the availability of compute resources. 

For the step 2, we estimate the number of missing objects per frame due to IoU deviation between bounding boxes detected at $f_{s(t-1)}$ and $f_{s(t)}$. To do so, we measure the number of objects $\tilde{m}_t$ during run-time that satisfy an IoU threshold between the two consequent detected frames
based on Algorithm~\ref{alg:riou measure}. 
\begin{algorithm}
\begin{algorithmic}
\STATE $\tilde{m}_t = 0$ \text{ // Initialize the number of survived objects.}  
\FOR{$i_o$ = 1, 2, ..., $\tilde{l}_{(t-1),c(t-1)}$}
\FOR{$j_o$ = 1, 2, ..., $\tilde{l}_{t,c(t)}$}
\STATE Measure IoU[$i_o$][$j_o$] between the two bounding boxes of $i_o$ and $j_o$
\IF {IoU[$i_o$][$j_o$] $\geq$ \text{IoU threshold}}
        \STATE $\tilde{m}_t = \tilde{m}_t + 1$
        \STATE \text{break }
\ENDIF  
\ENDFOR
\ENDFOR
\end{algorithmic}
\caption{Measuring the number of objects satisfying an IoU threshold}
\label{alg:riou measure}
\end{algorithm}
Using $\tilde{m}_t$, we seek the number of objects, $\bar{m}_t$, that violates an IoU threshold between the two detected frames as follows: 
\begin{equation}
   \bar{m}_t = \tilde{l}_{(t-1),c(t-1)} - \tilde{m}_t, 
\end{equation}
where $\tilde{l}_{(t-1),c(t-1)}$ is the number of detected objects at $f_{s(t-1)}$ measured by $d_{c(t-1)}$. Now, we can estimate the number of missing objects per frame, $u_t$, as follows: 
\begin{equation} \label{eq:u_t missing}
u_t = \bar{m}_t/b_{t,c(t)}.    
\end{equation}

For the step 3, we estimate the number of objects detected at $f_{s(t)+j}$, $q_{s(t)+j}$, using $u_t$ iteratively as follows:
\begin{equation} \label{eq:q_t n_est}
q_{s(t)+j} = q_{s(t)+j-1} - u_t,
\end{equation}
where $q_{s(t)} = \tilde{l}_{t,c(t)}$. Notice that we leverage temporal locality and assume that $u_t$ (measured between $f_{s(t-1)}$ and $f_{s(t)}$) can be applied for the frames from $f_{s(t)}$ to $f_{s(t+1)}$. The $u_t$ objects out of $q_{s(t)+j-1}$ objects at $f_{s(t)+j-1}$ are generally switched from TPs to FPs at $f_{s(t)+j}$, letting both precision and recall drop in proportion to the ratio of $(q_{s(t)+j} /q_{s(t)+j-1})$ at $f_{s(t)+j}$, compared to  $f_{s(t)+j-1}$. Therefore, it is highly probable that the AP at the  frame $f_{s(t)+j}$ drops quadradically in proportion to  $(q_{s(t)+j} /q_{s(t)+j-1})$, compared to the frame $f_{s(t)+j-1}$. Now, we estimate $\beta_{s(t)+j}$ based on this observation: 
\begin{equation} \label{eq:beta_eachframe}
\beta_{s(t)+j} = \beta_{s(t)+j-1} \times (q_{s(t)+j}/q_{s(t)+j-1})^2.
\end{equation} 

\subsection{Estimating Relative Average Precision} \label{sec:rap}
The RAP of $d_i$ to $d_{c(t)}$, $a_{t,i}$, can be expressed using an offline accuracy ratio between the two detectors, $\alpha_{t,i}$, and an accuracy degradation ratio between the two detectors, $\gamma_{t,i}$, as follows:
\begin{equation} \label{eq:rit}
a_{t,i} = \overline{AP}_{t,i}/\overline{AP}_{t,c(t)}  = \alpha_{t,i} \times \gamma_{t,i},
\end{equation}
where 
\begin{equation} \label{eq:alpha_i,t}
  \alpha_{t,i} = AP_{t,i}(f_{s(t)})/AP_{t,c(t)}(f_{s(t)}) \approx  l_{t,i}/\tilde{l}_{t,c(t)} 
\end{equation}
and 
\begin{equation} \label{eq:beta_i,t}
  \gamma_{t,i} = (\Sigma_{j=0}^{b_{t,i}-1} \beta_{s(t)+j}/b_{t,i})/(\Sigma_{j=0}^{b_{t,c(t)}-1} \beta_{s(t)+j}/b_{t,c(t)}).  
\end{equation}

ROMA chooses one of $d_i$s that has the index $i$ of the maximum $a_{t,i}$.  

\subsection{Implementation of ROMA}
This section exemplifies the implementation of ROMA (i.e., Eq.~(\ref{eq:rit})) in terms  of the initialization process and running process. 

\subsubsection{Initialization} At initialization time, {multiple detectors} are uploaded to DRAM. An FPS constraint, $FPS$, is found based on a video file. 
The initial frame block size $b_{i,0}$ uses the prior latency information of the detector $d_i$ on a compute platform. The histogram matrix $\mathbf{P}$ is found using a video dataset unseen from the evaluation dataset. The default detector is chosen as the slowest detector. The maximum frame block size is set to $30$, and the all $\beta_{s(0)+j}$s are initialized to `1' for $0 \leq j \leq 29$.  

\subsubsection{Running Process}
For the updates of ${\alpha}_{t,i}$ in Eq~(\ref{eq:alpha_i,t}), the $l_{t,i}$s are estimated using Eq.~(\ref{eq:l_it}).  To prevent the division by zero in Eq.~(\ref{eq:alpha_i,t}), we add $0.1$ to the divisor.
The $\mathbf{r}_{t,i}$ is found using Eq.~(\ref{eq:r_it}) and
the $\tilde{\mathbf{p}}_{t,c(t)}$ in Eq.~(\ref{eq:l_it}) is found using the detected bounding boxes information at the frame $f_{s(t)}$ using $d_{c(t)}$. Each frame block size $b_{t,i}$ is updated based on Eq.~(\ref{eq:b_it}). Depending on whether  $d_{c(t)}$ is changed between $f_{s(t-1)}$ and $f_{s(t)}$, each $L_{t,i}$ is estimated using Eq.~(\ref{eq:L_it1}) for $c(t) = c(t-1)$ or Eq.~(\ref{eq:L_it2}) and Eq.~(\ref{eq:L_it2_again}) for $c(t) \neq c(t-1)$.
The number of objects satisfying an IoU threshold is computed based on Algorithm~\ref{alg:riou measure} and the number of missing objects per frame $u_t$ is computed using Eq.~(\ref{eq:u_t missing}). Each $\beta_{s(t)+j}$ is computed based on Eq.~(\ref{eq:q_t n_est}) and Eq.~(\ref{eq:beta_eachframe}). 
The detector with the index $i$ that has the maximum value of ${a}_{t,i}$ in Eq.~(\ref{eq:rit}) is selected to be run at the frame $f_{s(t+1)}$. 

We also address a special case in which an $a_{t,i}$ is estimated using a faster detector $d_{c(t)}$ and a slower $d_i$. This case generates a lower $b_{t,c(t)}$ than  $b_{t,i}$. Since computing $\beta_{s(t)+j}$ relies on the current run-time information measured using $d_{c(t)}$, no run-time information is available for the $\beta_{s(t)+j}$ updates where $j \geq b_{t,c(t)}$. In this case, we leverage the ratio of $\beta_{s(t-1)+j}/\beta_{s(t-1)+j-1}$ for the $\beta_{s(t)+j}$ updates as follows: 
\begin{equation} \label{eq:beta_exception}
\beta_{s(t)+j} = \beta_{s(t)+j-1} \times \beta_{s(t-1)+j}/\beta_{s(t-1)+j-1}. 
\end{equation}
Eq.~(\ref{eq:beta_exception}) can update ${\beta}_{s(t)+j}$ up to the maximum frame block size depending on the detectors $d_i$s.  
We address another special case in which the accuracy of $\beta_{s(t)+j}$ can suffer from noise of bounding boxes severely when $b_{t,c(t)}$ is low. To mitigate its effect, we set up a minimum frame block size threshold to update $\beta_{s(t)+j}$: $b_{th} = 3$. \Eg, if a frame block size of a current detector is larger than or equal to $b_{th}$, we update $\beta_{s(t)+j}$. Otherwise, we utilize $\beta_{s(t-1)+j}$s for $\beta_{s(t)+j}$s.  

\section{Experimental Evaluation} \label{sec:experiment}
The experimental setting is as follows:
\\  - {Computing Platform:} An NVIDIA Jetson Nano Board (MAX power mode). \\  - {Object Detectors:} YOLOv4-Tiny-288 (YT288), YOLOv4-Tiny-416 (YT416), YOLOv4-Full-288 (YF288), and YOLOv4-Full-416 (YF416) optimized by TensorRT with an FP16 (i.e., half precision) option. The confidence score thresholds are set to 0.3 for all YOLOs. The IoU threshold in Algorithm~\ref{alg:riou measure} is set to 0.5.
\\  - {Evaluation Datasets:} MOT17Det and MOT20Det~\cite{MOTDet}.
\\  - {Prior Histogram Matrix $\mathbf{P}$:}
$\mathbf{P}$ with $H=3$, generated using four MOT15 datasets named ETH-Bahnhof, ETH-Sunnyday, TUD-Campus, and TUD-Stadtmitte, each having $640 \times 480$ resolutions \cite{leal-motchallenge15}. 
\begin{equation}
\mathbf{P}=
  \begin{bmatrix}
    1921 & 3550 & 2748 \\
    4603 & 3872 & 2488 \\
    8502 & 3506 & 2982 \\
    9526 & 3603 & 2993
  \end{bmatrix}
\end{equation}
We chose $H=3$, each for small object size region $r_s$, medium object size region $r_m$, and large object size region $r_l$. We set up the object size boundaries for $s_1 = 2500$ ($pixel^2$) between $r_s$ and $r_m$, and $s_2 = 7500$ between $r_m$ and $r_l$  with respect to a $640\times480$ resolution video frame so that each region can contain at least $20\%$ of detected objects out of total detected objects across the three regions. 
\\  - {Comparison with State-Of-The-Art Techniques:} YT288, YT416, YF288, YF416, TOD~\cite{lee-tod}, and LAD~\cite{lou-dynamic-ofa}. Notice that LAD in our paper utilizes the four YOLOv4 models instead of the detectors generated from \cite{cai-once}. It downgrades a detector to the next lighter detector (e.g., YT416 to YT288) when the latency violates an FPS constraint and upgrades a detector to the next heavier detector (e.g., YT288 to YT416) when the inference latency is lower than 30\% of the latency constraint. 
\\  - {Accuracy Evaluation Tool:} MATLAB interface MOT evaluation tool kit provided by \cite{MOTDet} (i.e., an 11-point interpolation assessment using an IoU threshold of $0.5$). If the precision is reported as `0' at the recall point `0' based on the evaluation tool, we take a precision at the recall point as: $p(r) = \underset{r'}{\mathrm{max}}\, p(r')$, where $p(r)$ is the precision value at a recall point $r$ and $r' \geq r$ \cite{manning-introduction}. 
\\ - {Real-Time AP:} We measure the real-time object detection accuracy as used in \cite{lee-tod, jiang-chameleon}; the bounding box information detected from the previous frame was used for the AP assessment for the subsequent dropped frames. 

\subsection{Real-Time AP Measurements}
We evaluate real-time APs for TOD, LAD, the four different YOLOv4s, and ROMA on MOT17/20Det datasets while imposing four different workloads as shown in  Table~\ref{tab:rtva_roma}: case (a) for no background workload, case (b) for background workload with running a YT288, case (c) for background workload with running a YT416, and case (d) for background workload with running a YF416. 

Using MOT17Det, ROMA outperforms all single detectors and other run-time techniques in terms of the average APs for each case of (a) to (d) as shown with bold marks in Table~\ref{tab:rtva_roma}. Table~\ref{tab:rtva_roma} shows that deploying one single resource-efficient detector limits the real-time accuracy for dynamically varying video contents and compute resources using MOT17Det (e.g., motivation of run-time techniques such as \cite{jiang-chameleon, lou-dynamic-ofa, lee-tod}). \Eg, deploying YF416 can be a good choice for case (a), but can be the worst choice for case (d).  

\begin{table*}
  \caption{Real-Time APs on MOT17/20Det with Background Workloads}
  \label{tab:rtva_roma}
  \centering
  \includegraphics[width=\linewidth]{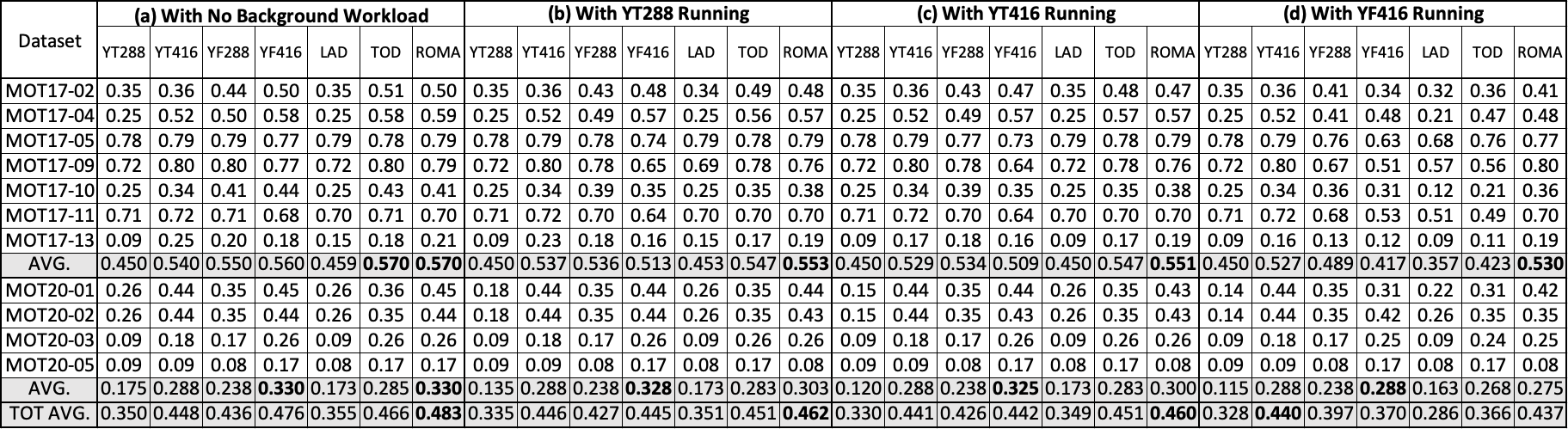}
\end{table*}

\begin{figure}[!t] 
\centering
\includegraphics[width=2.5in]{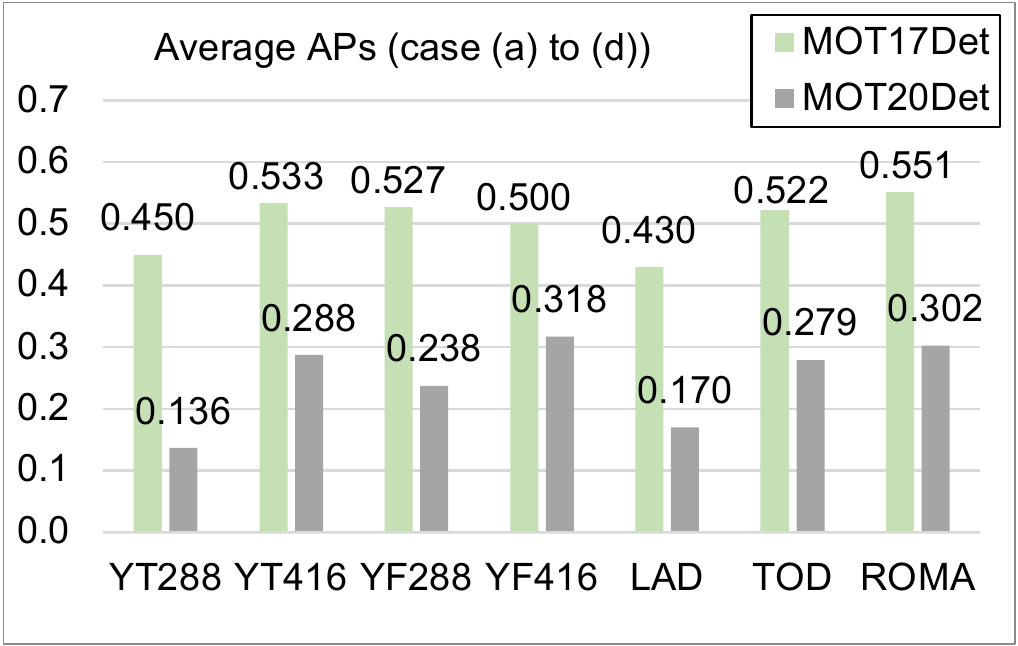}
\caption{Average APs (MOT17Det and MOT20Det)}
\label{fig:ap_mot17and20}
\end{figure}

Fig.~\ref{fig:ap_mot17and20} shows the average APs of all detectors across cases (a) to (d) using MOT17Det and MOT20Det, respectively. This scenario mimics a dynamically varying compute resources and video content scenario in which each of the four different compute resources (i.e., (a) to (d)) is available for $1/4$ of the entire execution time and each video dataset is included in proportion to the number of frames of the dataset. The ROMA shows the accuracy improvements of 1.23, 1.03, 1.05, 1.10, 1.28, and 1.06 $\times$ compared to YT288, YT416, YF288, YF416, LAD, and TOD, respectively using MOT17Det, even though there are effectively fewer valuable YOLOs to choose from. Therefore, the AP difference between ROMA and any individual YOLO can be limited. This implies that ROMA is suitable for dynamically varying compute resources for each different video content case. 

\begin{figure}[!t] 
\centering
\includegraphics[width=\linewidth]{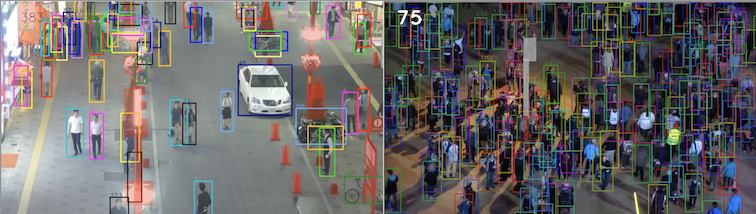}
\caption{MOT17-04 (Left) and MOT20-05 (Right) \cite{MOTDet}}
\label{fig:mot17_20_comp}
\end{figure}

For MOT20Det, ROMA is the second best detector, following YF416, since MOT20Det contains more people than MOT17Det. \Eg, Fig.~\ref{fig:mot17_20_comp} shows MOT17-04 and MOT20-05, respectively. The time overhead of ROMA 
quadratically increases in proportion to the number of detected objects as shown in Algorithm~\ref{alg:riou measure}. \Eg, the time overhead of ROMA on an NVIDIA Jetson Nano is measured as $6 ms$ for case (a) on MOT17-04 and $12 ms$ on MOT20-05. Notice that the time overhead does not depend on the number of objects on a video frame but on the number of detected objects using a detector.  YF416 is chosen by ROMA with 100\% for both MOT17-04 and MOT20-05 as shown in Fig.~\ref{fig:deployment_frequency}.  Considering the detection latency of YF416 ($225 ms$), the time overhead of ROMA did not affect the real-time accuracy on MOT17-04, but on MOT20-05 across case (a) to (d) based on Table~\ref{tab:rtva_roma}. However, ROMA has equivalent performance to YF416 on the other MOT20Det datasets. Even though TOD and LAD have lower time overhead on average  than ROMA (\eg, $0.5 ms$ for TOD and $0.01 ms$ for LAD for MOT17Det), the decision of ROMA is more accurate than TOD and LAD, resulting in higher real-time accuracy.  


\begin{figure}[!t] 
\centering
\includegraphics[width=2.5in]{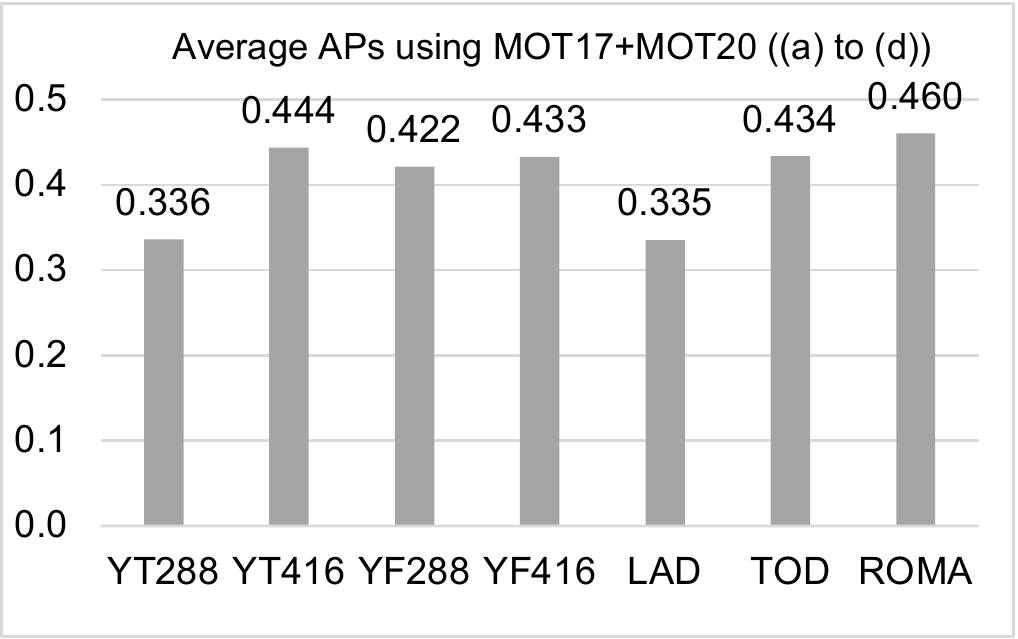}
\caption{Average APs across All Cases (MOT17Det+MOT20Det)}
\label{fig:avgmot1720}
\end{figure}

Finally, we consider another scenario containing both MOT17Det and MOT20Det to compute the average AP of each detector across the four cases. In this scenario, Fig.~\ref{fig:avgmot1720} shows that ROMA is the best performing detector, showing $1.37$, $1.04$, $1.09$, $1.06$, $1.37$, and $1.06 \times$ performance improvement, compared to YT288, YT416, YF288, YF416, LAD, and TOD, respectively.   

\subsection{Decisions by ROMA}
\begin{figure}[!t] 
\centering
\includegraphics[width=2.7in]{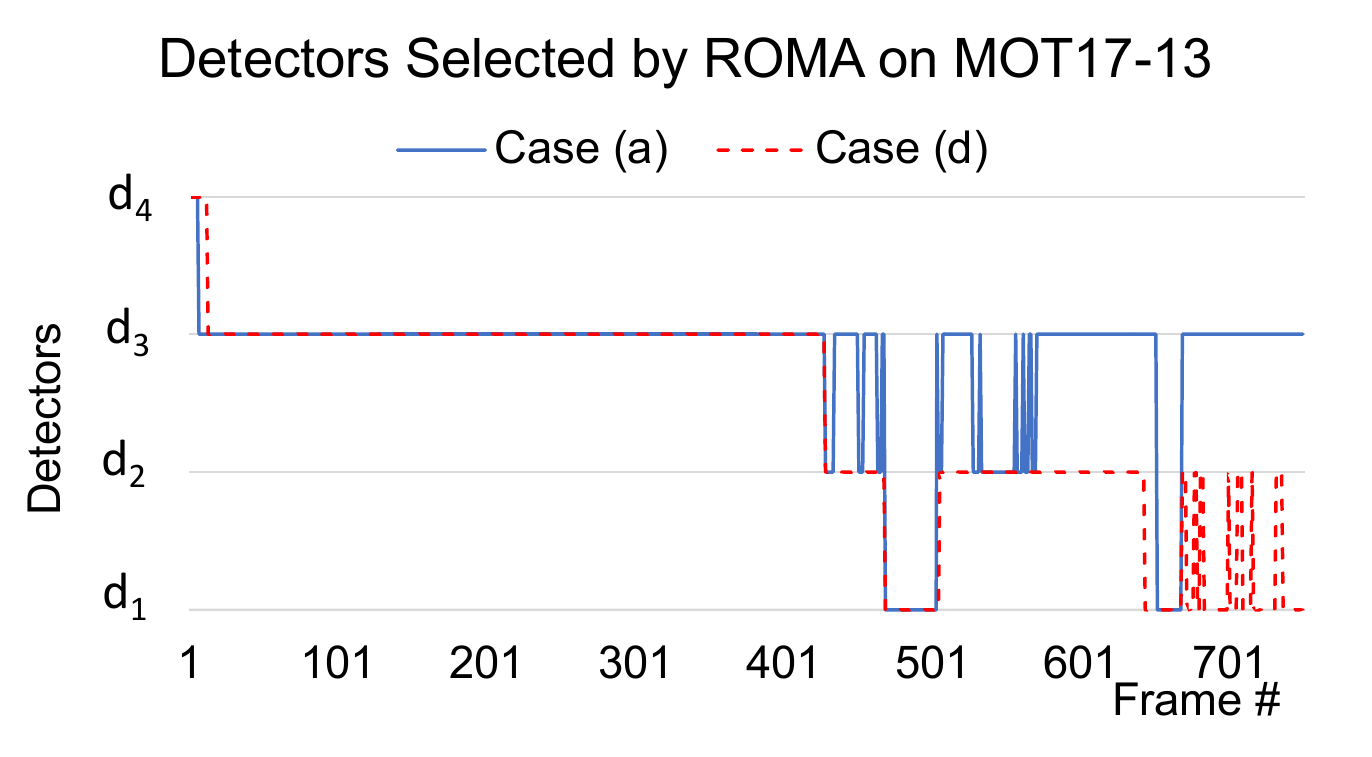}
\caption{Detectors selected by ROMA on MOT17-13}
\label{fig:sel_detectors_13}
\end{figure}

Fig.~\ref{fig:sel_detectors_13} shows the decisions made by ROMA with $d_1 = YT288$, $d_2 = YT416$, $d_3 = YF288$, and $d_4=YF416$ for case (a) and (d) on MOT17-13.ROMA downgrades the detectors used for case (a) to lighter detectors for case (d).
A bus moves straight in the direction aligned with the camera in early frames and turns right later, increasing relative object speeds. ROMA switches a current detector to a lighter detector when objects move faster. With a low dynamic range of object sizes and the object moving speeds, the decision of ROMA was biased in early frames.

\begin{figure}[!t] 
\centering
\includegraphics[width=\linewidth]{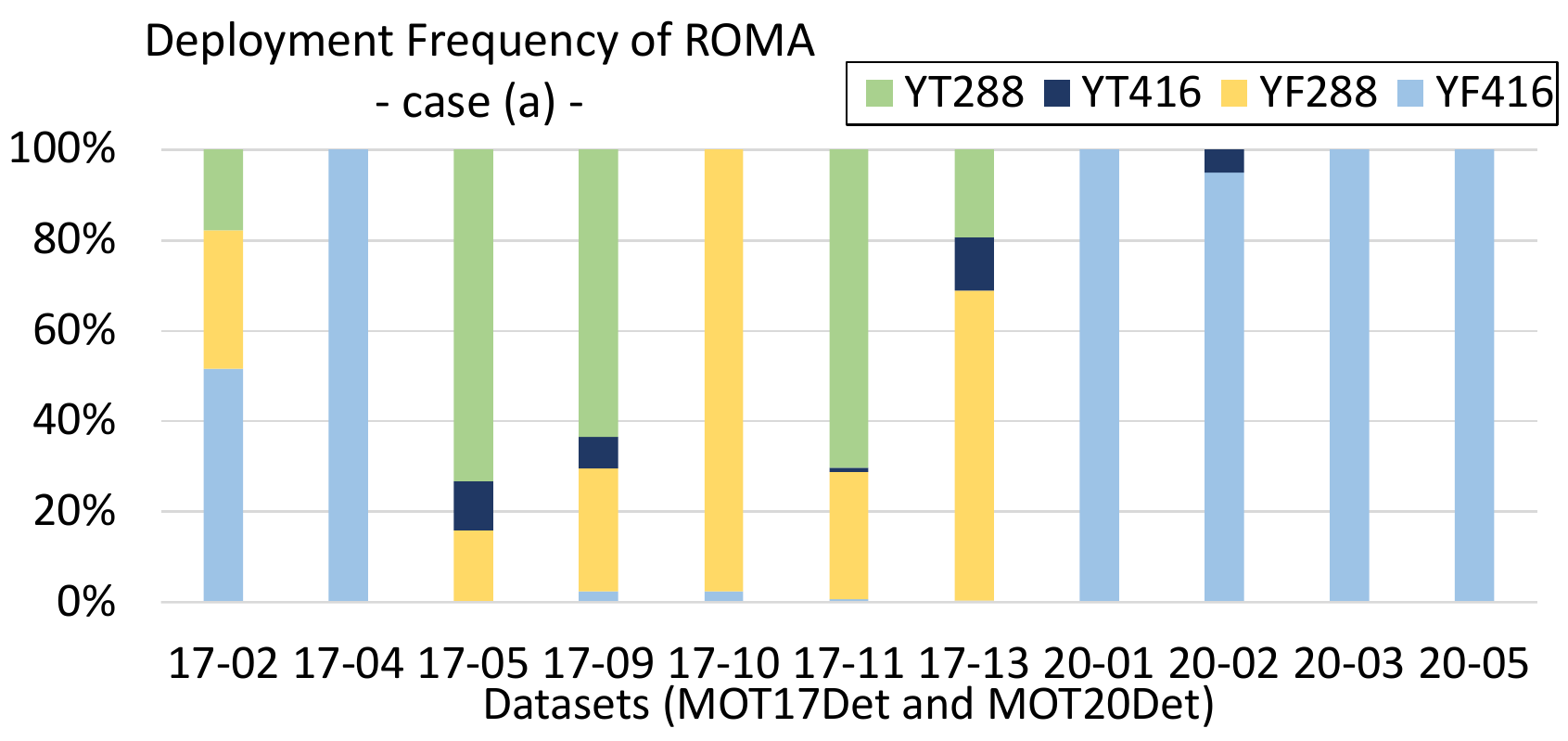}
\caption{Deployment Frequency by ROMA}
\label{fig:deployment_frequency}
\end{figure}

Fig.~\ref{fig:deployment_frequency} shows the deployment frequency of each detector by ROMA on MOT17Det and MOT20Det. ROMA selects YF416 with 100\% for MOT17-04, MOT20-01, MOT20-03, and MOT20-05 (\eg, video frames captured by static cameras), while selects multiple detectors dynamically for MOT17-05, 09, 11, and 13 (\eg, video frames captured by moving cameras). ROMA selects multiple detectors in MOT17-02 (static camera), since people walk in early frames and later kids riding bicycles appear, increasing relative object speeds. 

\section{Conclusion} \label{sec:conclusion}
Deploying a single object detector limits real-time accuracy on dynamically varying video contents and compute resources due to the fixed structure of the detector, which is the motivation of run-time techniques \cite{jiang-chameleon, lou-dynamic-ofa, lee-tod} and our paper. To our best knowledge, no literature had yet discussed how to select an appropriate detector without label information according to both dynamically varying video contents and available compute resources. ROMA is designed to be able to switch between multiple detectors without label information according to both dynamically varying video contents and available compute resources.

This paper claims that the run-time information including the object size histograms, the IoUs approximation, and the detection latency is sufficient to estimate relative APs accurately for all detector candidates according to dynamically varying video contents and compute resources. ROMA on an NVIDIA Jetson Nano demonstrates the best real-time accuracy in a scenario of dynamically varying video contents and available compute resources based on MOT17Det and MOT20Det, compared to individual YOLOv4 detectors and two state-of-the-art run-time techniques.

\end{document}